\documentclass{IOS-Book-Article}

\usepackage{mathptmx}
\usepackage[utf8]{inputenc}
\usepackage{graphicx, dblfloatfix}
\usepackage{tabularx}
\usepackage{soul}
\usepackage{times}
\usepackage{url}
\usepackage{latexsym}
\usepackage{layout}
\usepackage{float}
\usepackage{placeins}
\usepackage{wrapfig}
\usepackage{afterpage}
\usepackage[normalem]{ulem}
\usepackage{color}
\usepackage{nicefrac}
\usepackage{multirow}
\usepackage{todonotes}

\def\hb{\hbox to 10.7 cm{}}

\begin{document}

\pagestyle{headings}
\def\thepage{}

\newcommand{\XXX}[1]{\textcolor{red}{XXX #1}}
\def\repl#1#2{\textcolor{red}{XXX \sout{#1}}\textcolor{blue}{#2}}

\begin{frontmatter}              

\title{Impact of Corpora Quality on Neural Machine Translation}

\markboth{}{September 2018\hb}

\author{
    \fnms{Matīss} \snm{Rikters}\thanks{Corresponding Author: Matīss Rikters; E-mail: name.surname@tilde.lv.}
}

\runningauthor{M. Rikters}
\address{Tilde, Vienības gatve 75A, Rīga, Latvia}

\begin{abstract}
    Large parallel corpora that are automatically obtained from the web, documents or elsewhere often exhibit many corrupted parts that are bound to negatively affect the quality of the systems and models that learn from these corpora. This paper describes frequent problems found in data and such data affects neural machine translation systems, as well as how to identify and deal with them. The solutions are summarised in a set of scripts that remove problematic sentences from input corpora.
\end{abstract}

\end{frontmatter}
\markboth{September 2018\hb}{September 2018\hb}
\thispagestyle{empty}
\pagestyle{empty}

\section{Introduction}
Machine translation (MT) systems - both, statistical (SMT) and neural (NMT) -  rely on large amounts of parallel data for training the models. It is often the case that larger amounts of corpora lead to higher quality models, therefore a common practice is automatic extraction of such corpora from web resources, digitised books and other sources. Such data is prone to be noisy and include all kinds of problematic sentences alongside the high-quality ones. Data quality plays an important role in training of statistical and, especially, neural network based models like NMT, which is quick to memorise bad examples. In the case of training SMT and NMT systems, often the only pre-processing is done using scripts from the Moses Toolkit \cite{Koehn2007Moses:Translation}, which is only capable of removing sentences that are longer or shorter than a specified amount or the source-target length ratio is too high.

In this paper, we explore the types of low-quality sentences commonly found in parallel corpora. We also compare the benefits of using additional filters to remove these sentences before training MT systems in contrast to using only the Moses scripts. We introduce a set of corpora cleaning tools
\footnote{Corpora Cleaning Tools: \url{https://github.com/M4t1ss/parallel-corpora-tools} } 
that remove sentences that have some of the most common problems found in large corpora. It is published in GitHub with the MIT open-source license.

\section{Related Work}

Zipporah \cite{Xu2017} is a trainable tool for selecting a high-quality subset of data from a huge amount of noisy data. The authors report that it can improve MT quality by up to 2.1 BLEU, but in order to use it, the tool requires a known high-quality data set for training.

Wolk \cite{Wolk2015NOISY-PARALLELLEVEL} proposes a method that uses online MT engines to translate source sentences from a parallel corpus and compare them with the given target sentences. It is very expensive to use on real-world parallel corpora, containing tens of millions of parallel sentences. The author reports results on using the method on rather small corpora of only several million words.

Khadivi and Ney \cite{Khadivi2005AutomaticTranslation} introduce a parallel corpora filtering method based on word alignment models. Similar to Zipporah, this method also relies on training using a high-quality corpus.

\section{Problems in Corpora}
\label{sec:problems}

This section outlines some often occurring problems in parallel corpora. The specific examples were obtained from the English-Estonian part of the ParaCrawl\footnote{Large-Scale Parallel Web Crawl: \url{http://statmt.org/paracrawl}} corpus.

\begin{figure}[ht]
    \includegraphics[width=\linewidth]{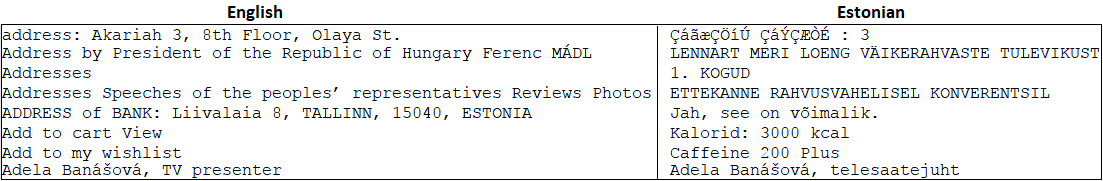}
    \caption{An example of a high mismatch in non-alphabetical character counts between source and target.}
    \label{fig:NonMatchingNonAlpha}
\end{figure}

\begin{figure}[ht]
  \includegraphics[width=\linewidth]{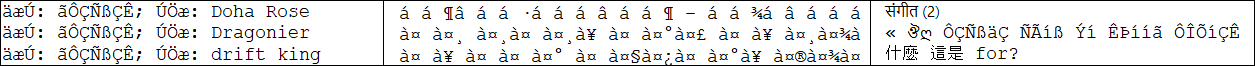}
  \caption{Examples of sentences with over 50\% non-alphabetical symbols.}
  \label{fig:NonAlpha}
\end{figure}

One of the most common defects in parallel corpora is a high mismatch between the non-alphabetic characters between source and target sentences (Figure \ref{fig:NonMatchingNonAlpha}). Also often are sentences that are completely or mostly composed of characters outside the scope of the language in question (Figure \ref{fig:NonAlpha}).

\begin{figure}[ht]
	\centering
  	\includegraphics[scale=0.6]{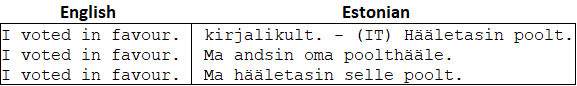}
    \caption{An example of an English sentence aligned to multiple different Estonian sentences.}
    \label{fig:OneToMultiple}
\end{figure}

\begin{figure}[ht]
	\centering
    \includegraphics[scale=0.6]{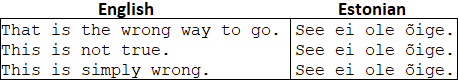}
    \caption{Multiple English paraphrased sentences aligned to one Estonian sentence.}
    \label{fig:ManyParaphrasesToOne}
\end{figure}

\begin{figure}[ht]
  \includegraphics[width=\linewidth]{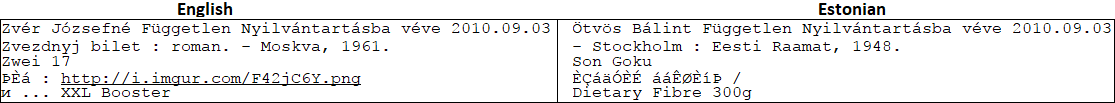}
  \caption{Examples of sentences with a different identified language than the one specified.}
  \label{fig:WrongLanguageIdentified}
\end{figure}

In parallel corpora, we may occasionally see the same sentence of one language aligned to multiple different ones of the other language (Figure \ref{fig:OneToMultiple}), but this is not always a bad indication, since they may just be paraphrases of the same concept (Figure \ref{fig:ManyParaphrasesToOne}). It is also wise to check if sentences in specific languages actually consist of text in that language (Figure \ref{fig:WrongLanguageIdentified}) as there may be citations and other parts of foreign language texts, especially in news domain corpora.

\begin{figure}[ht]
  \includegraphics[width=\linewidth]{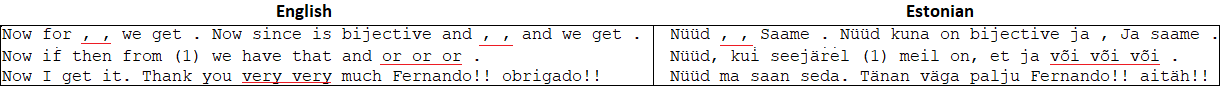}
  \caption{An example repeating tokens (underlined).}
  \label{fig:RepeatingTokens}
\end{figure}

Finally, a little less common observation for automatically gathered corpora, but somewhat more often in automatically generated (translated) parallel corpora is the repeating of tokens (Figure \ref{fig:RepeatingTokens}). Sentences like this may not always be incorrect, but they introduce ambiguity when used to train MT systems.

\section{Corpora Filters} 
\label{sec:filters}

The filters described in this section are mainly intended for parallel corpora consisting of two files with identical line-counts where each line of one file is related to the same line of the other file. Several of the filters are applicable to monolingual data as well and can be used to clean data for unsupervised MT training, back-translation, and other use-cases.

\textbf{Unique parallel sentence filter} -- removes duplicate source-target sentence pairs.

\textbf{Equal source-target filter} -- removes sentences that are identical in the source side and the target side of the corpus.

\textbf{Multiple sources - one target} and \textbf{multiple targets - one source} filters -- removes repeating sentence pairs where the same source sentence is aligned to multiple different target sentences and multiple source sentences aligned to the same target sentence.

\textbf{Non-alphabetical filters} -- remove sentences that contain over 50\% non-alphabetical symbols on either the source side or the target and sentence pairs that have significantly more (at least 1:3) non-alphabetical symbols in the source side than in the target side (or vice versa).

\textbf{Repeating token filter} -- especially useful for filtering back-translated parallel corpora that are created by translating a clean monolingual corpus into another language using NMT. NMT output may sometimes exhibit repeated words in the generated translation, which indicates that the system had problems translating a part of the sentence and it used the repetitions to fill the gap. In such cases the source-target sentence pair is likely to not be a good parallel sentence, therefore the repeating token filter removes them.

\textbf{Correct language filter} -- uses language identification software \cite{Lui2012Langid.py:Tool} to estimate the language of each sentence and removes any sentence that has a different identified language from the one specified.

\textbf{Moses Scripts and Subword NMT} -- calls Moses scripts for tokenising, cleaning, truecasing, and Subword NMT \cite{Sennrich2015} for splitting into subword units. This process prepares the corpus up to the point where it can be passed on to the NMT system for training.

\section{Experiments and Results} 
\label{sec:experiments}

\begin{table*}[h]
    \centering
    \caption{Detailed results on filtering English-Estonian/Finnish/Latvian larger common parallel corpora from WMT shared tasks.}
	\begin{tabular}{lllllllll}
	\hline
	 & \multicolumn{2}{c}{Paracrawl} & \multicolumn{3}{c}{Rapid} & \multicolumn{3}{c}{Europarl} \\  
	 & \multicolumn{1}{c}{En-Et} & \multicolumn{1}{c}{En-Fi} & \multicolumn{1}{c}{En-Et} & \multicolumn{1}{c}{En-Fi} & \multicolumn{1}{c}{En-Lv} & \multicolumn{1}{c}{En-Et} & \multicolumn{1}{c}{En-Fi} & \multicolumn{1}{c}{En-Lv} \\ \hline
	\multicolumn{1}{l}{Corpus size} & 1298103 & 624058 & 226978 & 583223 & 306588 & 652944 & 1926114 & 638789 \\ \hline
	\multicolumn{1}{l}{\multirow{2}{*}{Unique}} & 26 & 37 & 23 & 161463 & 80894 & 23218 & 52686 & 19652 \\  
	\multicolumn{1}{l}{} & 0.00\% & 0.01\% & 0.01\% & \textbf{27.68\%} & \textbf{26.39\%} & 3.56\% & 2.74\% & 3.08\% \\ \hline
	\multicolumn{1}{l}{\multirow{2}{*}{src == tgt}} & 242816 & 41611 & 428 & 3488 & 2929 & 490 & 528 & 707 \\
	\multicolumn{1}{l}{} & \textbf{18.71\%} & \textbf{6.67\%} & 0.19\% & 0.60\% & 0.96\% & 0.08\% & 0.03\% & 0.11\% \\ \hline
	\multicolumn{1}{l}{\multirow{2}{*}{\begin{tabular}[c]{@{}l@{}}* sources\\ 1 target\end{tabular}}} & 267235 & 17239 & 1108 & 1513 & 990 & 1176 & 6631 & 979 \\  
	\multicolumn{1}{l}{} & \textbf{20.59\%} & 2.76\% & 0.49\% & 0.26\% & 0.32\% & 0.18\% & 0.34\% & 0.15\% \\ \hline
	\multicolumn{1}{l}{\multirow{2}{*}{\begin{tabular}[c]{@{}l@{}}* targets\\ 1 source\end{tabular}}} & 69225 & 9532 & 752 & 1016 & 329 & 462 & 3536 & 435 \\  
	\multicolumn{1}{l}{} & \textbf{5.33\%} & 1.53\% & 0.33\% & 0.17\% & 0.11\% & 0.07\% & 0.18\% & 0.07\% \\ \hline
	\multicolumn{1}{l}{\multirow{2}{*}{\begin{tabular}[c]{@{}l@{}}\textgreater 50\%\\ non-alpha\end{tabular}}} & 200338 & 12919 & 1226 & 5647 & 1699 & 66 & 285 & 72 \\  
	\multicolumn{1}{l}{} & \textbf{15.43\%} & 2.07\% & 0.54\% & 0.97\% & 0.55\% & 0.01\% & 0.01\% & 0.01\% \\ \hline
	\multicolumn{1}{l}{\multirow{2}{*}{\begin{tabular}[c]{@{}l@{}}Non-alpha\\ mismatch\end{tabular}}} & 23777 & 12737 & 6674 & 13311 & 6361 & 7211 & 24847 & 4012 \\  
	\multicolumn{1}{l}{} & 1.83\% & 2.04\% & 2.94\% & 2.28\% & 2.07\% & 1.10\% & 1.29\% & 0.63\% \\ \hline
	\multicolumn{1}{l}{\multirow{2}{*}{\begin{tabular}[c]{@{}l@{}}Repeating\\ tokens\end{tabular}}} & 11210 & 1397 & 175 & 396 & 171 & 727 & 2594 & 703 \\  
	\multicolumn{1}{l}{} & 0.86\% & 0.22\% & 0.08\% & 0.07\% & 0.06\% & 0.11\% & 0.13\% & 0.11\% \\ \hline
	\multicolumn{1}{l}{\multirow{2}{*}{\begin{tabular}[c]{@{}l@{}}Language\\ mismatch\end{tabular}}} & 283152 & 36233 & 14762 & 24854 & 8739 & 8924 & 10932 & 3301 \\  
	\multicolumn{1}{l}{} & \textbf{21.81\%} & \textbf{5.81\%} & \textbf{6.50\%} & \textbf{4.26\%} & 2.85\% & 1.37\% & 0.57\% & 0.52\% \\ \hline
	\multicolumn{1}{l}{\multirow{2}{*}{$\sum$ removed}} & 1097779 & 131705 & 25148 & 211688 & 102112 & 42274 & 102039 & 29861 \\  
	\multicolumn{1}{l}{} & \textbf{85\%} & \textbf{21\%} & \textbf{11\%} & \textbf{36\%} & \textbf{33\%} & 6\% & 5\% & 5\% \\ \hline
	\end{tabular}
    \label{tab:full-results}
\end{table*}

\subsection{Corpora Cleaning}

We used the toolkit to clean parallel corpora provided in the WMT17\footnote{Second Conference on Machine Translation - \url{http://statmt.org/wmt17}} and WMT18\footnote{Third Conference on Machine Translation - \url{http://statmt.org/wmt18}} news MT shared tasks for English $\leftrightarrow$ Estonian/Finnish/Latvian. Detailed results of the cleaning process for three of the largest corpora - ParaCrawl, Rapid corpus of EU press releases (Rapid) and European Parliament Proceedings Parallel Corpus (Europarl) - are shown in Table \ref{tab:full-results}.

The results show that ParaCrawl is the most problematic corpus, especially the Estonian part, where 85\% had to be removed. The most frequent problems are 1) specified and identified language mismatch; 2) identical sentences appearing on source and target sides; 3) multiple source sentences aligned to the same target sentence; 4) an overwhelming amount of non-alphabetical characters; and 5) multiple target sentences aligned to the same source sentence. All examples of bad sentences in Section \ref{sec:problems} were selected from the removed parts of the English-Estonian ParaCrawl corpus.

The Rapid corpus had an overall higher quality with only about 25\% of parallel sentences removed. For the three languages it exhibited three main defects - 1) duplicate parallel sentences; 2) specified and identified language mismatch; and 3) mismatch in amounts of non-alphabetical symbols between source and target sentences.

Europarl was by far the cleanest corpus, having only 5-6\% of sentences removed by the cleaning toolkit. For all languages, most removed sentences were due to the same two defects as in the Rapid corpus.

We combined and shuffled all three English-Estonian corpora, resulting in 1 012 824 (46.50\% of total) sentence parallel corpus for training NMT systems described in the next section. The total amount of English-Finnish parallel sentences was 2 719 104 (82.72\% of total) after adding a cleaned version of the Wiki Headlines corpus, and English-Latvian - 1 617 793 (35.85\% of total) parallel sentences after adding cleaned versions of LETA translated news, Digital Corpus of European Parliament (DCEP), and Online Books corpora (cleaning details in Table \ref{tab:small-results}). We used the development data sets provided by the WMT shared tasks.

\begin{table}[h]
    \caption{Detailed results on filtering English-Finnish/Latvian smaller parallel corpora from WMT shared tasks.}
    \begin{tabular}{lllll}
    \hline
     & \multicolumn{1}{c}{En-Fi} & \multicolumn{3}{c}{En-Lv} \\
     & \multicolumn{1}{c}{Wiki} & \multicolumn{1}{c}{DCEP} & \multicolumn{1}{c}{Leta} & \multicolumn{1}{c}{Books} \\  \hline
    Corpus size & 153728 & 3542280 & 15671 & 9577 \\ \hline
    \multirow{2}{*}{Unique} & 0 & 2277397 & 454 & 434 \\ 
     & 0.00\% & \textbf{64.29\%} & 2.90\% & 4.53\% \\ \hline
    \multirow{2}{*}{src == tgt} & 42438 & 339861 & 2 & 4 \\ 
     & \textbf{27.61\%} & \textbf{9.59\%} & 0.01\% & 0.04\% \\ \hline
    \multirow{2}{*}{\begin{tabular}[c]{@{}l@{}}* sources\\ 1 target\end{tabular}} & 161 & 12474 & 2 & 35 \\ 
     & 0.10\% & 0.35\% & 0.01\% & 0.37\% \\ \hline
    \multirow{2}{*}{\begin{tabular}[c]{@{}l@{}}* targets\\ 1 source\end{tabular}} & 339 & 9450 & 15 & 12 \\ 
     & 0.22\% & 0.27\% & 0.10\% & 0.13\% \\ \hline
    \multirow{2}{*}{\begin{tabular}[c]{@{}l@{}}\textgreater 50\%\\ non-alpha\end{tabular}} & 488 & 31842 & 0 & 13 \\ 
     & 0.32\% & 0.90\% & 0.00\% & 0.14\% \\ \hline
    \multirow{2}{*}{\begin{tabular}[c]{@{}l@{}}Non-alpha\\ mismatch\end{tabular}} & 4616 & 38838 & 946 & 20 \\ 
     & 3.00\% & 1.10\% & 6.04\% & 0.21\% \\ \hline
    \multirow{2}{*}{\begin{tabular}[c]{@{}l@{}}Repeating\\ tokens\end{tabular}} & 38 & 1242 & 47 & 8 \\ 
     & 0.02\% & 0.04\% & 0.30\% & 0.08\% \\ \hline
    \multirow{2}{*}{\begin{tabular}[c]{@{}l@{}}Language\\ mismatch\end{tabular}} & 74507 & 48910 & 59 & 1074 \\ 
     & \textbf{48.47\%} & 1.38\% & 0.38\% & \textbf{11.21\%} \\ \hline
    \multirow{2}{*}{$\sum$ removed} & 122587 & 2760014 & 1525 & 1600 \\ 
     & \textbf{80\%} & \textbf{78\%} & \textbf{10\%} & \textbf{17\%} \\ \hline
    \end{tabular}
    \label{tab:small-results}
\end{table}

\subsection{Machine Translation}

To observe the actual benefit of filtering data for NMT, we trained NMT models using filtered and non-filtered data in both translation directions for the three language pairs. We used Sockeye \cite{Sockeye:17} to train transformer architecture models with 6 encoder and decoder layers, 8 transformer attention heads per layer, word embeddings and hidden layers of size 512, dropout of 0.2, shared subword unit vocabulary of 50 000 tokens, maximum sentence length of 128 symbols, and a batch size of 3072 words. All models were trained until they reached convergence on development data.

The final NMT system results in Table \ref{tab:mt-results} show that corpora filtering improves NMT quality for Estonian and Latvian systems, but not Finnish. The lack of improvement for Finnish is mainly due to the Europarl being the largest (about $\frac{3}{5}$ of total) and at the same time the cleanest corpus for this language pair. The biggest corpora for Estonian and Latvian - ParaCrawl (about $\frac{3}{5}$ of total) and DCEP (about $\frac{4}{5}$ of total) respectively were also the most problematic ones with 85\% and 78\% sentences removed respectively.

\begin{figure}[h]
  \includegraphics[width=\linewidth]{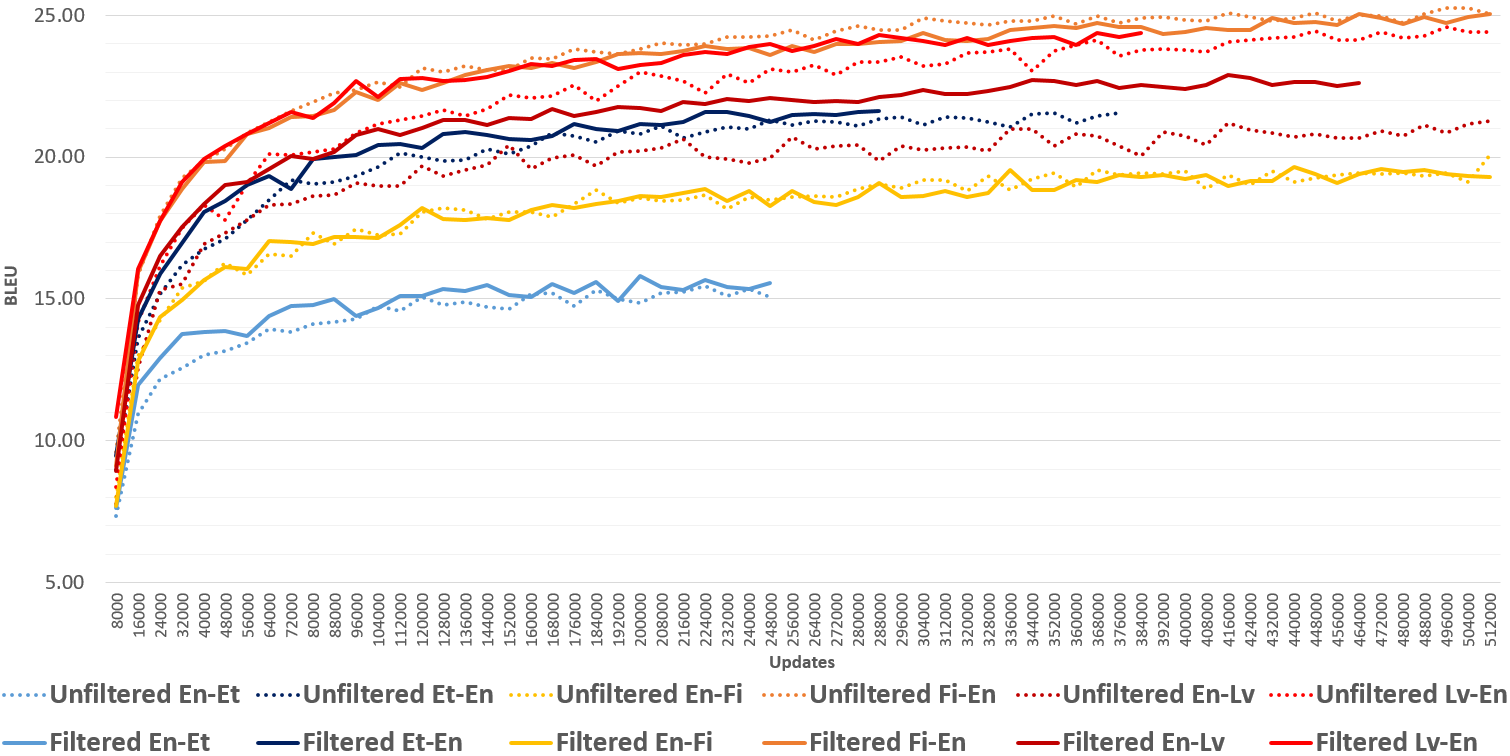}
  \caption{Training progress of English $\leftrightarrow$ Estonian/Finnish/Latvian NMT systems.}
  \label{fig:NMTProgress}
\end{figure}

Figure \ref{fig:NMTProgress} shows training progression of all 12 NMT systems. Filtered systems are depicted with solid lines, unfiltered ones - with dotted lines, Estonian systems are in light/dark blue colours, Finnish - orange/yellow, and Latvian are in light/dark red colours. The figure shows that the filtered Estonian and Latvian systems are much quicker to learn than the unfiltered ones, but eventually, they converge close to the unfiltered systems. As for the Finnish systems - there is no significant difference between filtered and unfiltered, as at times one is higher than the other or vice versa.

It is generally visible that in both translation directions the filtered systems achieve higher BLEU scores and reach higher quality quicker. For both English-Estonian systems, the unfiltered version catches up to the filtered one later on in the training, but never quite reaches or surpasses it. 

\begin{table}[h]
    \caption{Translation quality results (BLEU scores) for all translation directions on development data. The best results are marked in bold. The second row shows how much of the initial parallel corpora remained after filtering for each language pair.}
    \begin{tabular}{lllllll}
    \hline
     & En-Et & Et-En & En-Fi & Fi-En & En-Lv & Lv-En \\ \hline
    Unfiltered & 15.45 & 21.55 & \textbf{20.07} & \textbf{25.25} & 21.29 & 24.12 \\
    Corpus after filtering & \multicolumn{2}{c}{46.50\%} & \multicolumn{2}{c}{82.72\%} & \multicolumn{2}{c}{35.85\%} \\
    Filtered & \textbf{15.80} & \textbf{21.62} & 19.64 & 25.04 & \textbf{22.89} & \textbf{24.37} \\
    Difference & \textbf{+0.35} & \textbf{+0.07} & -0.43 & -0.21 & \textbf{+1.60} & \textbf{+0.25} \\ \hline
    \end{tabular}
    \label{tab:mt-results}
\end{table}

\section{Conclusion} 
\label{sec:conclusion}

This paper introduced several types of problematic sentences that can be found in large text corpora and a set of filters that help to remove them in order to train higher quality neural machine translation models using the remaining clean part of the corpora. Results show that in cases where the majority of given parallel corpora are very noisy and there is a small fraction of high-quality corpora, cleaning boosts NMT performance. This is especially evident for translation into morphologically rich languages like Estonian and Latvian.

In this paper, we mainly focused on cleaning parallel corpora, but the toolkit is also capable of cleaning monolingual corpora separately. In the MT system training workflow, cleaning monolingual data is useful before performing back-translation of an in-domain corpus, so that only filtered sentences get translated.

We release the corpora cleaning toolkit on GitHub under the MIT open-source license. The toolkit was used as an integral part of the runner-up English-Estonian NMT system submission \cite{pinnisetal2018} in the WMT18 news translation task for cleaning parallel and back-translatable monolingual data, as well as synthetic parallel data produced via back-translation.

\section*{Acknowledgements}
\label{sec:acknowledgments}

The research has been supported by the European Regional Development Fund within the research project ”Neural Network Modelling for Inflected Natural Languages” No. 1.1.1.1/16/A/215.

\label{sec:ref}

\bibliographystyle{ios1}
\bibliography{Mendeley}

\end{document}